\documentclass[runningheads]{llncs}
\usepackage{hyperref} 
\usepackage{amsmath}
\usepackage{amsfonts}
\usepackage{booktabs}
\usepackage{subfig}
\usepackage{pgfplots}
\usepackage{pgfplotstable}
\pgfplotsset{compat=1.18}
\usepackage[dvipsnames]{xcolor}

\usepackage[capitalize]{cleveref}
\crefname{section}{Sec.}{Secs.}
\Crefname{section}{Section}{Sections}
\Crefname{table}{Table}{Tables}
\crefname{table}{Tab.}{Tabs.}

\usepackage{tabularx}
\newcolumntype{C}{>{\centering\arraybackslash}X}

\usepackage[T1]{fontenc}
\usepackage{graphicx}
\makeatletter
\renewcommand{\subsubsection}{%
  \@startsection{subsubsection}{3}%
  {\z@}%
  {-8\p@ \@plus -4\p@ \@minus -4\p@}% {-18\p@ \@plus -4\p@ \@minus -4\p@}%
  {-0.5em \@plus -0.22em \@minus -0.1em}%
  {\normalfont\normalsize\bfseries\boldmath}%
}
\makeatother
\makeatletter
\renewcommand{\paragraph}{%
  \@startsection{paragraph}{4}%
  {\z@}{0.5em}{-0.5em}%
  {\normalfont\normalsize\itshape}%
}
\makeatother

\input{macros}
\begin{document}
\title{Towards Scalable Backpropagation-Free\\Gradient Estimation}
\titlerunning{Towards Scalable Backpropagation-Free Gradient Estimation}
%\author{Anonymous}
\author{Daniel Wang\orcidID{0009-0005-1349-1552} \and
Evan Markou\orcidID{0000-0001-8476-7310} \and
Dylan Campbell\orcidID{0000-0002-4717-6850}}
\authorrunning{D. Wang et al.}
\institute{Anonymous Institution}
\institute{Australian National University, Canberra, Australia\\
    \email{\{u7918232, Evan.Markou, Dylan.Campbell\}@anu.edu.au}
    }%\and

\maketitle % typeset the header of the contribution
\begin{abstract}
While backpropagation---reverse-mode automatic differen\-tiation---has been extraordinarily successful in deep learning, it requires two passes (forward and backward) through the neural network and the storage of intermediate activations.
Existing gradient estimation methods that instead use forward-mode automatic differentiation struggle to scale beyond small networks due to the high variance of the estimates.
Efforts to mitigate this have so far introduced significant bias to the estimates, reducing their utility.
We introduce a gradient estimation approach that reduces both bias and variance by manipulating upstream Jacobian matrices when computing guess directions.
It shows promising results and has the potential to scale to larger networks, indeed performing better as the network width is increased.
Our understanding of this method is facilitated by analyses of bias and variance, and their connection to the low-dimensional structure of neural network gradients.

\end{abstract}
\section{Introduction}
While backpropagation \cite{rumelhart1986learning} is ubiquitous in deep learning, there has been recent interest in alternative memory-efficient methods of optimising a neural network without using backpropagation \cite{singhal2023,baydin2022,silver2022learning,chandra2021unexpected}. 
One such class of methods is based on estimating the gradient using forward-mode automatic differentiation (AD).
Forward-mode AD efficiently evaluates the Jacobian--vector product in a single forward pass and avoids the memory cost of storing intermediate activations required by backpropagation.
Additional motivation for backpropagation alternatives arises from the fact that backpropagation is biologically implausible \cite{crick1989}.

Using forward-mode AD, an unbiased estimator of the gradient can be constructed by randomly sampling a guess direction and scaling it by the directional derivative in the guess direction.
This method is defined as the forward gradient \cite{ren2023}.
However, because this estimator requires sampling in a high-dimensional guessing space equal to the number of parameters, high variance prohibits this method from scaling beyond training small networks on toy datasets.
For a network with $N$ parameters, the cosine similarity between the estimated and true gradient is $O(\frac{1}{\sqrt{N}})$, which means that the guess direction becomes orthogonal to the true gradient as network size increases.
A number of works based on the forward gradient, including state-of-the-art gradient guessing method ``$\tilde{W}\transpose$'' \cite{singhal2023}, have focused on decreasing its variance by reducing the guessing space
% and generating better guess directions that align with the true gradient
\cite{ren2023,fournier2023,singhal2023}.
% Most notably, the current state-of-the-art gradient guessing method ``$W\transpose$'' \cite{singhal2023} serves as both the foundation of the methods that are introduced in this project and the baseline to which they will be evaluated against.
Neural network gradients have also been shown to exhibit a low-dimensional structure \cite{gurari2018,song2025}, which if properly exploited has the potential to significantly improve gradient guessing methods.

% In this paper, we introduce two novel methods based on $W\transpose$.
% By manipulating the upstream Jacobian matrices, the \textit{Preconditioning} method corrects the bias of $W\transpose$ at the cost of higher variance, and the \textit{$W^\perp$} method lowers both bias and variance.
% Experiments show at the $W^\perp$ method performs better than the Preconditioning method and achieves notable improvement over $W\transpose$ for larger networks.
In this paper, we introduce a novel method $\tilde{W}^\perp$ that extends the $\tilde{W}\transpose$ approach, reducing both its bias and its variance by considering the low-rank structure of the guessing space.
Our experiments show that the $\tilde{W}^\perp$ method performs considerably better than $\tilde{W}\transpose$ for larger networks.
We also observe a possible connection to the low-dimensional structure of the gradient in that it lies in a low-dimensional subspace of the image of the upstream Jacobian matrix, providing additional insight into our gradient guessing methods.  

\section{Related Work and Technical Background}

\subsubsection{Forward-mode Automatic Differentiation (AD).}
Given a function $f:\mathbb{R}^n \rightarrow \mathbb{R}^m$, forward-mode AD computes the Jacobian--vector product $Jv$ where $J \in \mathbb{R}^{m\times n}$ is the Jacobian matrix and $v \in \mathbb{R}^n$ is an arbitrary vector. 
In the context of optimising a neural network where the loss function $L:\mathbb{R}^n \rightarrow \mathbb{R}$ typically has a scalar output, the Jacobian--vector product is simply the directional derivative in the direction $v$. 
% In practice, automatic differentiation frameworks such as PyTorch \cite{paszke2019pytorch} and JAX \cite{jax2018github} provide functions to compute this efficiently.
In practice, automatic differentiation frameworks such as PyTorch and JAX provide functions to compute this efficiently.

\subsubsection{Forward Gradient (Weight Perturbation \cite{baydin2022}).}
For the remainder of this paper we will consider an $l$-layer MLP with weights $W$ and biases $b$ with layer widths of $d_1, d_2, \ldots, d_l$, ReLU activations, and loss function $L$. 
The $i$th linear layer can be written as
\begin{equation}
s_i = W_i x_i + b_i, \quad s_i \in \mathbb{R}^{d_{i+1}}\,,
\end{equation}
where $W_{i} \in \mathbb{R}^{d_{i+1} \times d_{i}}$, $b_i \in \mathbb{R}^{d_{i+1}}$, and
\begin{equation}
x_{i+1} = \mathrm{ReLU}(s_i), \quad x_{i+1} \in \mathbb{R}^{d_{i+1}}\,.
\end{equation}
We will only consider optimising the weights $W_i$, but these methods can be easily extended to the biases too. 
At layer $i$, the vanilla forward gradient method generates a guess of
$\partial L / \partial W_i$
as follows.
First, a guess direction $y_i$ of the same dimension as $W_i$ is sampled from the standard normal distribution $\mathcal{N}(\mathbf{0}, \mathbf{I})$.
Then, the Jacobian--vector product $y\transpose \frac{\partial L}{\partial W}$---the directional derivative of $\partial L / \partial W$ in the direction $y$, where $W$ denotes the concatenated $W_i$s and $y$ denotes the concatenated $y_i$s---is computed in the forward pass. 
The guess for $\partial L / \partial W_i$ is defined as
$\hat{\frac{\partial L}{\partial W_i}} = (y_i\transpose\frac{\partial L}{\partial W_i})y_i$:
the guess direction scaled by the directional derivative in that direction.
With this approach, weight updates are performed by replacing the true gradient with the guess $\hat{\partial L} / \partial W_i$. 
% This method is also referred to as \textit{weight perturbation}. 
In general, sampling the guess direction from a zero-mean unit-variance distribution with independent components yields an unbiased but high variance estimator of the true gradient.

\subsubsection{Activation perturbation \cite{ren2023}.}
This method guesses the gradient with respect to the activations instead of the weights. 
In deep neural networks, the weight matrix $W_i \in \mathbb{R}^{d_{i+1}\times d_i}$ contains many more entries than the activation vector $s_i \in \mathbb{R}^{d_{i+1}}$. 
Therefore, guessing the gradient with respect to the activations $\partial L / \partial s_i$ reduces the dimension of guessing space and hence decreases variance. 
Similar to before, the guess
$\hat{\frac{\partial L}{\partial s_i}} = (y\transpose \frac{\partial L}{\partial s})y_i$,
where $y$ denotes the concatenated guess directions $y_i$ and $s$ denotes the concatenated activations $s_i$ at each layer,
is computed by sampling a random vector $y_i \in \mathbb{R}^{d_{i+1}}$ and scaling it by the Jacobian--vector product $y\transpose \frac{\partial L}{\partial s}$. 
Given the guess $\hat{\frac{\partial L}{\partial s_i}}$, the gradient guess with respect to the weight matrix can be obtained via the chain rule,
\begin{equation}
\hat{\frac{\partial L}{\partial W_i}} = \hat{\frac{\partial L}{\partial s_i}}  \frac{\partial s_i}{\partial W_i} = \hat{\frac{\partial L}{\partial s_i}} x_i\transpose\,.
\end{equation} 
As well as reducing variance, Ren \etal \cite{ren2023} show that activation perturbation is also unbiased.
Consequently, it performs better than weight perturbation.

%\textbf{Gradient guessing with local losses}. An extension of the forward gradient is also proposed in \cite{ren2023}, where the network is partitioned into submodules each equipped with its own local loss function. Each submodule estimates only the gradient of its local loss function with respect to the parameters of that submodule via activation perturbation. By estimating the gradient of each submodule separately, this method achieves significant variance reduction and performs the best among their proposed methods. \cite{fournier2023} consider a variation of this approach, where instead of estimating the gradient of each local loss, they explicitly compute the gradients of the local losses via backpropagation and use them as a guess direction for the global loss function gradient. The idea is that the deterministic local gradients approximate the global gradient better than randomly sampled guess directions, and as a result this method achieves better performance than random guess directions. Because backpropagation is only performed locally, this is much more efficient than full backpropagation. However, they also observe that the local and global gradients are weakly aligned which fundamentally limits the effectiveness of local loss methods. 

\subsubsection{$\tilde{W}\transpose$ method \cite{singhal2023}.}
The $\tilde{W}\transpose$ method extends the activation perturbation approach by incorporating terms along the chain rule gradient decomposition up to the next layer's activations into the guess for $\partial L / \partial s_i$.
For a network with linear layers and ReLU activations, the chain rule gives  
\begin{equation}
\hat{\frac{\partial L}{\partial W_i}} = \hat{\frac{\partial L}{\partial s_{i+1}}} \frac{\partial s_{i+1}}{\partial x_{i+1}} \frac{\partial x_{i+1}}{\partial s_i} \frac{\partial s_i}{\partial W_i} = \epsilon_{i+1}\transpose W_{i+1} M_i x_i\transpose\,,
\label{eq4}
\end{equation}
where $\epsilon_{i+1}$ is a sample from a standard normal distribution that approximates the unknown $\partial L / \partial s_{i+1}$, $W_{i+1}$ is the weight matrix of the next layer, $M_i$ is the diagonal matrix corresponding to the ReLU derivative mask, and $x_i$ is the input to layer $i$.
Using the first three terms in \cref{eq4}, a guess direction $y_i\transpose = \epsilon_{i+1}\transpose  W_{i+1} M_i$ for $\partial L / \partial s_i$ is generated by sampling $\epsilon_{i+1}$ and multiplying with the next layer weight matrix $W_{i+1}\transpose$ and the ReLU mask $M_i$.
As before, the Jacobian--vector product is $\frac{\partial L}{\partial s}^\top y$, and for each layer the guess for $\frac{\partial L}{\partial s_i}$ is $\hat{\frac{\partial L}{\partial s_i}} = (y\transpose \frac{\partial L}{\partial s}) y_i$.
The weight update for $W_i$ is obtained from the last term of \cref{eq4} by taking the outer product $\hat{\frac{\partial L}{\partial s_i}} x_i\transpose$.

The intuition behind this approach is that it leverages information from the network architecture and narrows the guessing space with each additional term in the chain rule. 
The next layer weight matrix is typically low-rank and the ReLU mask is sparse, leading to a large variance reduction.
Although we focus on an MLP with ReLU activations, in a more general setting the guess direction can be simply written as $y_i = \tilde{W}_{i+1}\transpose \epsilon_{i+1}$, where $\tilde{W}_{i+1}$ represents the composition of each term in the chain rule from the current layer's activations $s_i$ up to the next layer's activations $s_{i+1}$.
For a ReLU MLP, $\tilde{W}_{i+1} = W_{i+1} M_i$.

However, because $y_i=\tilde{W}_{i+1}\transpose \epsilon_{i+1}$ is no longer a zero-mean unit-variance random variable with independent components, this estimator is no longer unbiased.
Singhal \etal \cite{singhal2023} show that the bias at layer $i$ is given by 
\begin{equation}
\mathbb{E}\left[\hat{\frac{\partial L}{\partial s_i}}\right] - \frac{\partial L}{\partial s_i} = (\mathrm{Cov}(y_i)-I)\frac{\partial L}{\partial s_i}\,,
\label{bias}
\end{equation}
where $\partial L / \partial s_i$ is the true gradient and $y_i$ is the guess direction with 
\begin{equation} 
\mathrm{Cov}(y_i) = \tilde{W}_{i+1}\transpose \tilde{W}_{i+1}\,.
\label{cov}
\end{equation}
Despite this incurred bias, the $\tilde{W}\transpose$ method significantly outperforms activation perturbation due to a large decrease in variance.

\subsubsection{Gradients lies in a low-dimensional subspace.}
The gradient guessing methods described above are based on the principle of reducing variance by reducing the dimensionality of the guessing space.
Related to the idea of gradients within a low-dimensional subspace is the work of Gur-Ari \etal \cite{gurari2018}, which observes that for a deep neural network trained on a $k$-class classification task, the gradient predominantly lies in the low-dimensional subspace spanned by the eigenvectors corresponding to the top-$k$ largest eigenvalues of the Hessian.  
Interestingly, Song \etal \cite{song2025} found that performing SGD in the small top-$k$ subspace fails to decrease the loss, but performing SGD in the orthogonal complement of this subspace recovers the original SGD performance, leading to the hypothesis that this phenomenon is caused by noise in stochastic gradient descent as opposed to full-batch gradient descent. 
In the context of forward gradient guessing methods, this phenomenon reveals the underlying structure of gradients which can be exploited by, as we show in the next section, guessing within a small subspace that aligns with the true gradient.

% \section{Method}
% \section{Reducing the Bias and Variance of the Guess Direction}
\section{A Low-Bias, Low-Variance Guess Direction}

We propose the $\tilde{W}^\perp$ method, based on the $\tilde{W}\transpose$ method, that reduces both the bias and variance of the guess direction by orthogonalising the upstream Jacobian matrix.
A variant, $\tilde{W}^\perp$-NS, accelerates the orthogonalisation using Newton--Schulz iterations.
Finally, to better understand the effect of bias and variance on the performance of gradient guessing methods, we also introduce a preconditioning method for comparison that is unbiased but has higher variance.

\subsection{$\tilde{W}^\perp$: Orthogonalising the Guess Projection Matrix}

In this section, we consider layer $i$ in isolation and drop the layer index $i$, so the general upstream Jacobian matrix up to the next layer is denoted by $\tilde{W}\transpose$ and the guess direction is $y = \tilde{W}\transpose \epsilon$.
That is, $\tilde{W}\transpose$ projects the Gaussian sample $\epsilon$ to the guess direction $y$.
Recall that for a ReLU MLP, $\tilde{W}_{i+1} = W_{i+1} M_i$, where $W_{i+1}$ is the weight matrix of the next layer and $M_i$ is the ReLU derivative mask, a diagonal binary matrix.
The central idea is to manipulate $\tilde{W}$ into an orthonormal, low-rank matrix $\tilde{W}'$ that further narrows the guessing space while bringing the covariance matrix $\mathrm{Cov}(y) = \tilde{W}^{'\mathsf{T}} \tilde{W}'$ closer to the identity, thus reducing both bias and variance.
We expect $\mathrm{rank}(\tilde{W})$ to be relatively low, since for a ReLU network it is at most the number of non-zero activations.
Let the reduced SVD of $\tilde{W}$ be
\begin{equation}
\tilde{W} = U_r\Lambda_r V_r\transpose\,,
\end{equation}
where $r=\mathrm{rank}(\tilde{W})$.
Then, we manipulate $\tilde{W}$ into the orthonormal matrix 
\begin{equation}
\tilde{W}'=U_rV_r\transpose\,,
\end{equation} so the guess direction is
$y = \tilde{W}^{'\mathsf{T}} \epsilon$
with covariance matrix Cov($y$) = $\tilde{W}^{'\mathsf{T}} \tilde{W}' = V_rU_r\transpose U_rV_r\transpose = I_r$, being the identity matrix in the first $r$ diagonal elements and 0 everywhere else.
To control the bias--variance trade-off, integer values of $k$ from 1 to $r$ can be chosen as a hyperparameter to select only the rows and columns in $U_r$ and $V_r\transpose$ corresponding to the $k$ largest singular values so that $\tilde{W}_k'=U_kV_k\transpose$.
Smaller values of $k$ reduce the dimensionality of the guessing space at the cost of pushing the covariance matrix $I_k$ further from the identity and thus incurring higher bias.

\subsection{$\tilde{W}^\perp$-NS: Fast Orthogonalisation via Newton--Schulz Iterations}

The Newton--Schulz iteration method \cite{kovarik1970,Björck1971,jordan2024muon}
can be used instead of SVD to significantly speed up our $\tilde{W}^\perp$ method.
The idea is that, given an odd polynomial, applying it to a matrix $A=USV\transpose$ is the same as applying it  element-wise to the singular values $S$. That is, 
\begin{equation}
p(A) = \sum_{k=0}^{K} a_k A(A\transpose A)^k = U\left( \sum_{k=0}^{K} a_k S^{2k+1} \right)V\transpose = Up(S)V\transpose\,.
\end{equation}
Applying this $N$ times yields
\begin{equation}
p^N(A) = Up^N(S)V\transpose\,,
\end{equation} 
where on the left hand side $p^N(A)$ is the matrix polynomial applied $N$ times to $A$ and on the right hand side it is the real-valued polynomial applied $N$ times element-wise to the singular values.
With carefully chosen coefficients, repeated composition of the polynomial $p^N(A)$ can be made to converge to various functions as $N \rightarrow \infty$. 
Jordan \etal \cite{jordan2024muon} use this method to orthogonalise $p^N(A) \approx UV\transpose$ by mapping all singular values to 1.
They find a degree five polynomial with $N=5$ to  approximate the function $f(x) = 1$ on $[0, 1]$. 

The Newton--Schulz iteration can be easily adapted to our $\tilde{W}^\perp$ method, labelled $\tilde{W}^\perp$-NS.
We wish to obtain $p^N(\tilde{W}) \approx U_kV_k\transpose$ by mapping the singular values larger than $\sigma_k$ to 1 and the ones smaller than $\sigma_k$ to 0, where $\sigma_k \in [0, 1]$ is the $k$\textsuperscript{th} largest singular value.  
That is, we need to approximate the step function defined by $f_{\sigma_k}(x) = 0$ for $x \in [0, \sigma_k]$ and $f(x) = 1$ for $x \in (\sigma_k, 1]$, where $\sigma_k$ is constrained to lie in $[0, 1]$ by normalising $\tilde{W}$ to have unit spectral norm.
Since the step function is slightly more complex than $f(x)=1$ on [0, 1], we fit a degree seven polynomial with coefficients obtained via gradient descent.
% This variation of $\tilde{W}^\perp$ is called $\tilde{W}^\perp$-NS.

\subsection{$\tilde{W}^\text{P}$: Preconditioning the Guess Projection Matrix}

Both the $\tilde{W}\transpose$ and $\tilde{W}^\perp$ methods reduce variance and increase accuracy over the unbiased activation perturbation approach, but at the cost of introducing bias. 
In this section, we develop an unbiased estimator, based on the $\tilde{W}\transpose$ method, at the cost of increasing the variance.
This allows us to investigate the bias--variance trade-off.
% Continuing with the idea of manipulating the upstream Jacobian matrix, our preconditioning method recovers an unbiased estimator from the $\tilde{W}\transpose$ method.
Specifically, we pre-multiply $\tilde{W}\transpose$ by $(\tilde{W}\transpose \tilde{W})^{-1/2}$ so that $\tilde{W}^{'\mathsf{T}} \approx(\tilde{W}\transpose \tilde{W})^{-1/2} \tilde{W}\transpose$.
Then $\tilde{W}^{'\mathsf{T}} \tilde{W}' = I$, and $y=\tilde{W}^{'\mathsf{T}} \epsilon$ has identity covariance. 
To be precise, let the SVD of $\tilde{W}$ be $\tilde{W}=USV\transpose$.
Since $\tilde{W}$ is not guaranteed to be full-rank, we introduce a small constant $\sigma=10^{-5}$ and modify $\tilde{W}$ to be $\tilde{W}_\sigma = U(S^2+\sigma I)^{1/2}V\transpose$, so that 
\begin{equation}
    \tilde{W}_\sigma\transpose \tilde{W}_\sigma = V(S^2+\sigma I)V\transpose\,,
\end{equation} 
can be diagonalised while keeping $\tilde{W}_\sigma$ close to $\tilde{W}$.
Setting
\begin{equation}
\tilde{W}_\sigma^{'\mathsf{T}} = (\tilde{W}_\sigma\transpose \tilde{W}_\sigma)^{-1/2} \tilde{W}_\sigma\transpose\,,
\end{equation}
yields an unbiased estimator, because the covariance matrix is identity:
\begin{align}
\mathrm{Cov}(y) &= \tilde{W}_\sigma^{'\mathsf{T}} \tilde{W}_\sigma' \\
&= (\tilde{W}_\sigma\transpose \tilde{W}_\sigma)^{-1/2}\tilde{W}_\sigma\transpose \tilde{W}_\sigma(\tilde{W}_\sigma\transpose \tilde{W}_\sigma)^{-1/2} \\
&= V(S^2+\sigma I)^{-1/2}V\transpose V(S^2+\sigma I)V\transpose V(S^2+\sigma I)^{-1/2}V\transpose\\
&= I\,.
\end{align}
This method has high variance because $(\tilde{W}_\sigma\transpose \tilde{W}_\sigma)^{-1/2}$ contains large singular values, due to $\sigma$ being small.
Moreover, unlike the $W\transpose$ and $W^\perp$ methods, the dimension of the guessing space is not reduced, because $\tilde{W}_\sigma'$ is a full rank matrix.

\section{Experiments}

\subsection{Experiment Setup} 

\subsubsection{Implementation Details and Dataset.}
Following Singhal \etal \cite{singhal2023}, the experiments in this section involve optimising an MLP with three hidden layers of size 128 and ReLU activations.
Our methods and existing methods from previous works are implemented in PyTorch, using \texttt{functorch.jvp} for forward mode automatic differentiation \footnote{The code is available at \href{https://github.com/danielwang0452/Forward-W-Perp}{https://github.com/danielwang0452/Forward-W-Perp.}}.
The experiments were run on an Apple M1 Pro Macbook with an 8-core CPU and
14-core GPU with 16GB of unified memory on the MNIST-1D dataset \cite{greydanus2024} for 300 epochs.
Note that 300 epochs is not enough for convergence.
The choice of optimiser, AdamW \cite{loshchilov2019} with learning rate $10^{-4}$ and batch size 512, is taken from the best performing configuration in Singhal \etal \cite{singhal2023} with no additional hyperparameter tuning. 

\subsubsection{Metrics: Bias.}

% We report several metrics specific to the bias and variance, which are defined here. 

% \paragraph{Bias.}
For a gradient guess direction $y_i^b = {\tilde{W}_{i+1}}^{b\mathsf{T}} \epsilon_i^b$ at layer $i$ where $b$ indexes over a batch of size $B$, the bias magnitude is given by  
\begin{equation}
\text{Bias}\left(\hat{\frac{\partial L}{\partial s_i}}\right) = \Bigl\| \frac{1}{B}\sum_{b=1}^B(\mathrm{Cov}(y_i^b)-I)\frac{\partial L}{\partial s_i^b} \Bigr\|_2 \,. 
\label{bias_metric}
\end{equation}

% \paragraph{Variance.}
\subsubsection{Metrics: Variance.}
We compute the variance magnitude at layer $i$ via the decomposition of MSE into bias and variance using the empirical mean squared error (MSE) and squared bias, given by
\begin{equation}
\widehat{\text{Var}}\left(\hat{\frac{\partial L}{\partial s_i}}\right) =
\Bigl\|
\underbrace{
\frac{1}{B}\sum_{b=1}^B\left(\frac{\partial L}{\partial s_i^b} - \hat{\frac{\partial L}{\partial s_i^b}}\right)^{\circ 2}
}_{\text{MSE}}
-
\underbrace{
\left(\frac{1}{B}\sum_{b=1}^B\left(\mathrm{Cov}(y_i^b)-I\right)\frac{\partial L}{\partial s_i^b}\right)^{\circ 2}
}_{\text{Squared Bias}}
\Bigr\|_2 \,,
\label{variance}
\end{equation}
where $\circ 2$ is the element-wise square.

\iffalse
We compute the variance at layer $i$ using the empirical mean squared error (MSE) and bias.
The MSE is approximated with a batch of $B$ data points with one sample of $\epsilon_i$ for each data point:
\begin{equation}
    \widehat{\text{MSE}}\left(\hat{\frac{\partial L}{\partial s_i}}\right) = \frac{1}{B}\sum_{b=1}^B \left(\frac{\partial L}{\partial s_i^b} - \hat{\frac{\partial L}{\partial s_i^b}}\right)^2\,.
\end{equation}
We obtain the variance via the decomposition of MSE into bias and variance
\begin{equation}
    \widehat{\text{Var}}\left(\hat{\frac{\partial L}{\partial s_i}}\right) = \widehat{\text{MSE}}\left(\hat{\frac{\partial L}{\partial s_i}}\right) - \text{Bias}\left(\hat{\frac{\partial L}{\partial s_i}}\right)^2\,,
    \label{variance_old}
\end{equation}
and similarly quantify it by taking the L2 norm and averaging over the batch dimension.
\fi

% \paragraph{Covariance Frobenius Norm.}
\subsubsection{Metrics: Covariance Frobenius Norm.}

Another way to measure bias is to consider $\mathrm{Cov}(y_i)-I$ independent of the true gradient $\frac{\partial L}{\partial s}$.
% We quantify $(\mathrm{Cov}(y)-I)$ by taking the Frobenius norm 
% \begin{equation}
%    \text{Covariance Frobenius Norm} := ||(\mathrm{Cov}(y)-I)||_F\,,
% \end{equation}
We quantify this with the Frobenius norm $||(\mathrm{Cov}(y)-I)||_F$,
averaged over the batch dimension. Intuitively, a smaller Covariance Frobenius Norm means that the covariance matrix is closer to identity, resulting in lower bias.
%\begin{equation}
%   ||(\mathrm{Cov}(y)-I)||_F = \sqrt{\frac{1}{B d_id_{i+1}}\sum_{b=1}^B\sum_{m=1}^{d_i}\sum_{n=1}^{d_{i+1}}(\mathrm{Cov}(y)-I)_{bmn}^2}
%\end{equation}\\
%\textbf{Variance: Covariance Trace.}
%Similarly, the variance can be measured independent of $\frac{\partial L}{\partial s}$ by the trace of the covariance matrix averaged over the batch and matrix dimensions so that it is invariant to changes in batch size and network width. 
%\begin{equation}
%\mathrm{Tr(Cov(y))} = \frac{1}{B d_i}\sum_{b=1}^B\sum_{m=1}^{d_i}\mathrm{Cov}%(y)_{bmm}
%\end{equation}
%These metrics are recorded per-layer in the experiments. 

\subsection{Results}

\subsubsection{$W^\perp$: Optimal Choice of $k$.}

\begin{table}[t!]
\centering
\caption{Training accuracies after 300 epochs for different values of $k$ in the $\tilde{W}^\perp$ method.
The $\Tilde{W}^\perp$ accuracies are comparable to $\Tilde{W}\transpose$, with smaller values $k=10$ and $k=25$ reaching the highest accuracy. 
%than $W\transpose$ for to larger variance reductionwith smaller values $k=10$ and $k=25$ exceeding $W\transpose$ method.
In general, the trend is that smaller $k$ perform better due to larger variance reduction, despite increased bias.}
\label{tab:orthogonal_table}
\begin{tabular}{l l c}
\toprule
\textbf{Method} & $k$ & \textbf{Accuracy} \\
\midrule
$\tilde{W}\transpose$ &           & 0.312 \\
\midrule
$\tilde{W}^\perp$ & $\mathrm{rank}(\tilde{W})$ & 0.311 \\
$\tilde{W}^\perp$ & 25             & 0.323 \\
$\tilde{W}^\perp$ & 10             & \textbf{0.324} \\
$\tilde{W}^\perp$ & 1              & 0.275 \\
\bottomrule
\end{tabular}
\label{W_perp}
\end{table}

We evaluate $W^\perp$ with different values of $k$, which controls the bias--variance trade-off, against the baseline $W\transpose$ method to examine its effect on training accuracy, shown in \cref{tab:orthogonal_table}. 
Smaller values of $k$ reduce variance by reducing the dimensionality of the guessing space.
However, reducing $k$ also increases the bias since the covariance matrix $I_k$ is further from the identity. 
For $k=\text{rank}(W)$ the final training accuracy is almost identical to $W\transpose$, and as smaller values of $k$ are chosen the accuracy increases.
This result shows that a lower-dimensional guessing space with reduced variance counteracts the additional bias, except for the extreme low variance, high bias $k=1$ case where there is a sharp decrease in accuracy. 

\subsubsection{Why Small Values of $k$ Perform Best.}

The phenomenon that gradients lie in a low-dimensional subspace offers an explanation as to why small values of $k$ attain the highest accuracy. 
Given the idea behind the $\tilde{W}\transpose$ method, that multiplying by the upstream Jacobian matrix $\tilde{W}\transpose$ constrains the gradient guess to lie in its image, denoted by Im$(\tilde{W}\transpose)$, it is reasonable to speculate that an analogous phenomenon occurs here where the true gradient lies not only in Im$(\tilde{W}\transpose)$ but in a low-dimensional subspace of Im($\tilde{W}\transpose)$. 
This can be empirically verified by projecting the true gradient $\partial L / \partial s$ onto Im($\tilde{W}$), similar to Gurari \etal \cite{gurari2018}. 
In the reduced SVD of $\tilde{W} = U_k\Lambda_k V_k\transpose$, $U_k$ contains an orthonormal basis of Im$(\tilde{W}\transpose)$, and the overlap $o_\text{top}$ of the true gradient in Im($\tilde{W}\transpose$) can be computed by projecting $\frac{\partial L}{\partial s}$ onto Im($\tilde{W}\transpose$), given by
\begin{equation}
o_\text{top} = \frac{\left\|U_k U_k\transpose \frac{\partial L}{\partial s}\right\|}{\left\|\frac{\partial L}{\partial s}\right\|}\,.
\end{equation}
% where 
% \begin{equation}
% \frac{\partial L}{\partial s}_\text{top} = U_k U_k\transpose \frac{\partial L}{\partial s} \,.
% \end{equation}
\begin{figure}[t!]
\centering

\begin{minipage}{0.49\textwidth}
\centering
\begin{tikzpicture}
\begin{axis}[
    width=\textwidth,
    height=5.5cm,
    xlabel={Number of singular values $k$},
    ylabel={Overlap},
    title={Layer Width = 128},
    legend style={
        at={(0.575,0.4)},
        draw=none,
        anchor=north,
        legend columns=1
        },
    legend cell align={left},
    grid=major,
    ymin=0, ymax=1,
    xtick distance=20,
    xticklabel style={/pgf/number format/fixed, /pgf/number format/precision=0}
]

\addplot[thick, color=blue, no markers] table[x index=0, y index=1, col sep=comma] {overlap_data_128.csv};
\addlegendentry{Layer 1, rank = $64$}

\addplot[thick, color=red, no markers] table[x index=0, y index=2, col sep=comma] {overlap_data_128.csv};
\addlegendentry{Layer 2, rank = $97$}

\addplot[thick, color=teal, no markers] table[x index=0, y index=3, col sep=comma] {overlap_data_128.csv};
\addlegendentry{Layer 3, rank = $10$}

\addplot[dotted, thick, black] coordinates {(10,0) (10,1)};
\node[above] at (axis cs:10,1) {\scriptsize $k=10$};

\end{axis}
\end{tikzpicture}
\end{minipage}%
\hfill
\begin{minipage}{0.49\textwidth}
\centering
\begin{tikzpicture}
\begin{axis}[
    width=\textwidth,
    height=5.5cm,
    xlabel={Number of singular values $k$},
    ylabel={Overlap},
    title={Layer Width = 512},
    legend style={
        at={(0.56,0.4)},
        draw=none,
        anchor=north,
        legend columns=1},
    legend cell align={left},
    grid=major,
    ymin=0, ymax=1,
    xtick distance=50,
    xticklabel style={/pgf/number format/fixed, /pgf/number format/precision=0}
]

\addplot[thick, color=blue, no markers] table[x index=0, y index=1, col sep=comma] {overlap_data_512.csv};
\addlegendentry{Layer 1, rank = $260$}

\addplot[thick, color=red, no markers] table[x index=0, y index=2, col sep=comma] {overlap_data_512.csv};
\addlegendentry{Layer 2, rank = $387$}

\addplot[thick, color=teal, no markers] table[x index=0, y index=3, col sep=comma] {overlap_data_512.csv};
\addlegendentry{Layer 3, rank = $10$}

\addplot[dotted, thick, black] coordinates {(10,0) (10,1)};
\node[above] at (axis cs:10,1) {\scriptsize $k=10$};

\end{axis}
\end{tikzpicture}
\end{minipage}

\caption{
Overlap of the true gradient $\frac{\partial L}{\partial s}$ onto the $k$-dimensional subspace of $\mathrm{Im}({\Tilde{W}}\transpose)$ corresponding to the $k$ largest singular values of $W$, for layer widths of 128 and 512.
In both cases, $\frac{\partial L}{\partial s}$ predominantly lies in a subspace of much lower dimension than $\mathrm{rank}(W)$; around $k=10$ captures most of the gradient, reducing the guessing space while introducing minimal bias.
Note that the Layer 3 weight matrix is low rank since it is connected to the output layer of width 10.
}
\label{overlap}
\end{figure}
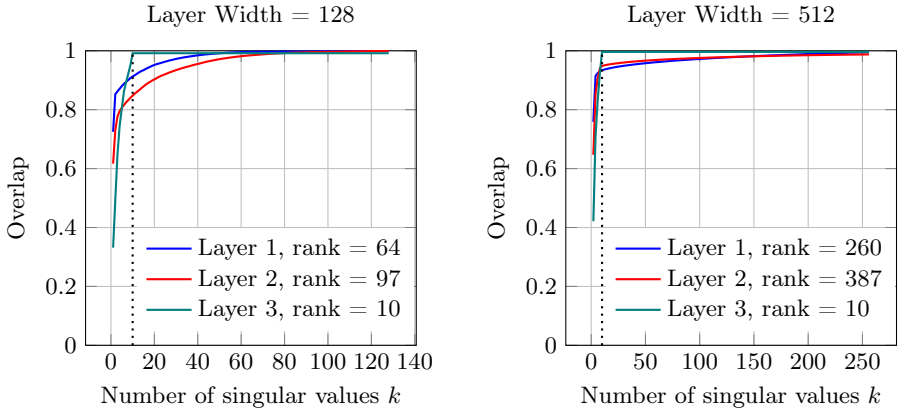

\cref{overlap} shows that the true gradient does in fact mostly lie in in the small $k$-dimensional subspace spanned by $V_k$.
Intuitively, choosing the smallest value of $k$ that is just large enough for the $k$-dimensional subspace to contain most of the true gradient will incur minimal additional bias while greatly reducing variance by narrowing the guessing space.
For the network architecture and dataset in our case, $k=10$ is a good choice.
\begin{table}[t!]
\centering
\caption{Variance, bias, and covariance Frobenius norm for different methods.}
\label{tab:cov_frobenius}
\begin{tabularx}{\linewidth}{l C C C}
% \begin{tabular}{lccc}
\toprule
\textbf{Method} & \textbf{Variance} \eqref{variance} & \textbf{Bias} \eqref{bias_metric} & \textbf{Cov. Norm}  \\
\midrule
$\tilde{W}^\top$               & $1.50 \times 10^{-4}$ & $2.75 \times 10^{-4}$ & \textbf{0.084} \\
$\tilde{W}^\perp$-bottom    & $1.30 \times 10^{-5}$ & $2.83 \times 10^{-4}$ & 0.085 \\
$\tilde{W}^\perp$ ($k=10$) & $\mathbf{1.03 \times 10^{-5}}$ & $\mathbf{2.86 \times 10^{-5}}$ & 0.085 \\
\bottomrule
\end{tabularx}
\label{orthogonal_bias_var}
\end{table}

\subsubsection{$\tilde{W}^\perp$: Understanding the Bias Reduction.}

Next, we observe that for appropriately chosen $k$, our $\tilde{W}^\perp$ method outperforms $\tilde{W}\transpose$ in both variance \textit{and} bias.
This is shown in \cref{orthogonal_bias_var}. 
The variance reduction is achieved by constraining the guesses to a low-dimensional subspace, while the bias reduction is achieved by aligning the guesses with the true gradient, which is less obvious. 
Recall that the bias is zero when the covariance matrix is identity. 
For $\tilde{W}^\perp$, the purpose of orthogonalisation is to zero out off-diagonal entries in the covariance matrix $I_k$ to bring it closer to identity. 
However, for small values of $k$, $I_k$ ends up being equally distant from identity (in the Frobenius norm) as $\tilde{W}\transpose \tilde{W}$, the covariance matrix of the $\tilde{W}\transpose$ method.
This might suggest that we have not actually decreased the bias in $\tilde{W}^\perp$. 
However, \cref{orthogonal_bias_var} shows that this is not the case. 

To illustrate that the bias reduction comes from aligning the guessing space with the true gradient, consider the following variant of $\tilde{W}^\perp$ which we refer to as $\tilde{W}^\perp$-bottom.
In the SVD of $\tilde{W}=USV\transpose$, we take $\tilde{W}'=U_kV_k\transpose$ corresponding to the $k=10$ \textit{smallest} (in this case 0) singular values, where we know the true gradient does not lie.
% (in contrast to $\tilde{W}^\perp$ where we take $\tilde{W}'=U_kV_k\transpose$ corresponding to the $k=10$ largest singular values where most of the true gradient lies). 
Both $\tilde{W}^\perp$ and $\tilde{W}^\perp$-bottom have the same covariance matrix $I_k$, so according to \cref{bias}, the difference in bias comes from the difference in the true gradients $\frac{\partial L}{\partial s}$. 
In other words, the optimisation trajectory of $\tilde{W}^\perp$ has smaller gradient norm $\left\| \frac{\partial L}{\partial s}\right\|_2$ than that of $\tilde{W}^\perp$-bottom, and hence lower bias.
The conclusion is that $\tilde{W}^\perp$ reduces bias by aligning the guessing space with the underlying low-rank structure of the true gradient. 

\subsubsection{$\tilde{W}^\perp$: Scaling to Larger Networks.}

\begin{figure}[t!]
  \centering
  \begin{tikzpicture}
    \begin{axis}[
        width=\textwidth,  % Full width
        height=6cm,
        xlabel={Layer Width},
        ylabel={Train Accuracy (300 Epochs)},
        ymin=0, ymax=1.05,
        xtick={64,128,256,512},
        xticklabel style={rotate=0},
        legend style={at={(0.5,-0.25)}, anchor=north, legend columns=3},
        legend cell align={left},
        grid=major,
        ymajorgrids=true,
        enlargelimits=0.1
    ]
    \addplot+[mark=star, thick, ForestGreen] table[x=Width, y={Weight Perturbation}, col sep=comma] {scaling_data.csv};
    \addlegendentry{Weight Perturbation}
    
    \addplot+[mark=square, thick, blue] table[x=Width, y={Activation Perturbation}, col sep=comma] {scaling_data.csv};
    \addlegendentry{Activation Perturbation}

    \addplot+[mark=o, thick, red] table[x=Width, y={W-Transpose}, col sep=comma] {scaling_data.csv};
    \addlegendentry{$\tilde{W}\transpose$}

    \addplot+[mark=triangle, thick, orange] table[x=Width, y={Orthogonal-WT}, col sep=comma] {scaling_data.csv};
    \addlegendentry{$\tilde{W}^\perp$ (k=10)}

    \addplot+[mark=diamond, thick, black] table[x=Width, y={Backprop}, col sep=comma] {scaling_data.csv};
    \addlegendentry{Backpropagation}

    \end{axis}
  \end{tikzpicture}
  \caption{Training accuracy as the hidden layer width increases.}
  \label{fig:scaling_plot}
\end{figure}
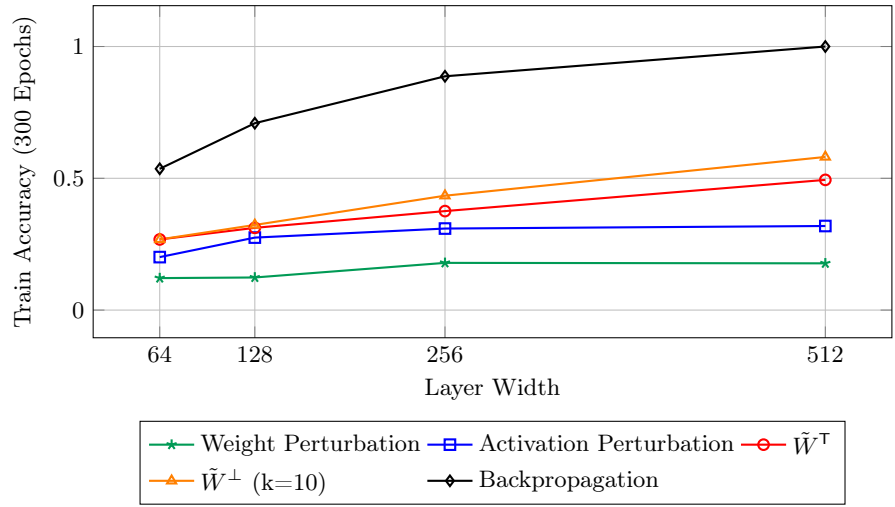

Compared to previous methods, the $\tilde{W}^\perp$ method shows promising results for larger networks. 
\cref{fig:scaling_plot} and \cref{tab:scaling_table} show that the gap in training accuracy between $\tilde{W}^\perp$ with $k=10$ and $\tilde{W}\transpose$ increases as the hidden layer width increases from 64 to 128 and 512. 
This behaviour can be explained by the variance reduction from narrowing the guessing space.
Recall the idea behind $\tilde{W}\transpose$, that multiplying a randomly sampled $\epsilon$ with $\tilde{W}\transpose$ restricts the guess direction $y=\tilde{W}\transpose\epsilon$ to lie in a lower dimensional Im($\tilde{W}\transpose$).
However, as layer width increases, so does the rank of the next layer weight matrix and hence the dimensionality of Im($\tilde{W}\transpose$). $\tilde{W}^\perp$ does not suffer from this problem, because the dimensionality of the guessing space can be controlled through the hyperparameter $k$. 
In our experiments, the value of $k=10$ which was observed to work well in the width 128 experiments still performs well for width 512 because the 10-dimensional subspace still contains a large portion of the true gradient. 
In general, these results indicate that for a layer of width $n$, most of  the true gradient will be contained in a $k$-dimensional subspace with $k<<n$, and that $k$ grows much slower than $n$. 
In practice, this means that the guessing space and hence variance of the $\tilde{W}^\perp$ method can be kept almost constant as layer width increases, enabling $\tilde{W}^\perp$ to be scaled to larger networks than previous methods.

% \vspace{1em}

\begin{table}[!t]\centering
\caption{
Train accuracies across methods and hidden layer widths. As layer width increases from 64 to 512, the accuracy gap between $\tilde{W}\transpose$ and $W^\perp$ increases from 0\% up to almost 10\%. This result indicates that $\tilde{W}^\perp$ will scale to larger networks much better than previous methods. However, there is still a significant gap between gradient guessing methods and backpropagation.
}
\resizebox{\textwidth}{!}{%
\begin{filecontents*}{scaling_data.csv}
Width, Weight Perturbation, Activation Perturbation,W-Transpose,Orthogonal-WT,Backprop
64, 0.121319115, 0.201, 0.268, 0.268, 0.536
128,0.1236854, 0.275,0.312,0.323,0.709
256,0.17908654, 0.30934495,0.3755601043,0.4339256897,0.887019217
512, 0.17734, 0.319,0.494,0.581,1.0
\end{filecontents*}
\pgfplotstabletypeset[
    col sep=comma,
    columns={Width,Weight Perturbation, Activation Perturbation,W-Transpose,Orthogonal-WT,Backprop},
    columns/Width/.style={
        column name=\shortstack{Layer\\Width},
        column type={>{\centering\arraybackslash}p{1.5cm}}
    },
    columns/Weight Perturbation/.style={
        column name=\shortstack{Weight\\Perturbation},
        fixed, fixed zerofill, precision=3,
        column type={>{\centering\arraybackslash}p{2.5cm}}
    },
    columns/Activation Perturbation/.style={
        column name=\shortstack{Activation\\Perturbation},
        fixed, fixed zerofill, precision=3,
        column type={>{\centering\arraybackslash}p{2.5cm}}
    },
    columns/W-Transpose/.style={
        column name=\shortstack{$\tilde{W}\transpose$},
        fixed, fixed zerofill, precision=3,
        column type={>{\centering\arraybackslash}p{2.3cm}}
    },
    columns/Orthogonal-WT/.style={
        column name=\shortstack{$\tilde{W}^\perp$\\(k=10)},
        fixed, fixed zerofill, precision=3,
        column type={>{\centering\arraybackslash}p{2.3cm}}
    },
    columns/Backprop/.style={
        column name=Backprop,
        fixed, fixed zerofill, precision=3,
        column type={>{\centering\arraybackslash}p{2.0cm}}
    },
    every head row/.style={before row=\toprule, after row=\midrule},
    every last row/.style={after row=\bottomrule}
]{scaling_data.csv}
}
\label{tab:scaling_table}
\end{table}

\begin{table}[t!]
\centering
\setlength{\tabcolsep}{10pt} % default is 6pt
\caption{Train and test accuracies and training times (300 epochs) for our unoptimised$^\dagger$ implementations of different methods across layer sizes of 128 and 512.
${}^\dagger$PyTorch MPS backend on MacOS laptop; SVD on the CPU.
}
\label{ns_table}
\begin{tabularx}{\linewidth}{l C C C C} % Added one more 'C' for the new column
\toprule
\textbf{Method} & \textbf{Layer Size} & \textbf{Train Time} & \textbf{Train Accuracy} & \textbf{Test Accuracy} \\ % Added "Time" column header
\midrule
$\tilde{W}^\perp$-NS      & 512 &   6h     & \textbf{0.594} & 0.477 \\
$\tilde{W}^\perp$, $k=10$  & 512 &   40h    & 0.581 & \textbf{0.498} \\
$\tilde{W}^\top$          & 512 &     26m    & 0.486 & 0.438 \\
\midrule
$\tilde{W}^\perp$-NS      & 128 &    9m & \textbf{0.326} & \textbf{0.291} \\
$\tilde{W}^\perp$, $k=10$  & 128 &  1h    & 0.324 & 0.283 \\
$\tilde{W}^\top$          & 128 &    5m       & 0.312 & 0.277 \\
\bottomrule
\end{tabularx}
\end{table} 
\subsubsection{$\tilde{W}^\perp$-NS Results.}

Here, we verify that the $\tilde{W}^\perp$-NS approximation to $\tilde{W}^\perp$ with $k=10$ does not harm performance. 
First, we describe a few implementation details that further speed up $\tilde{W}^\perp$-NS. 
To avoid having to repeatedly determine the polynomial coefficients that approximate $f_{\sigma_k}$ (since $\sigma_k$ varies across layers and over the course of training), we pre-determine a set of coefficients that approximate $f_{\sigma_k}$ at intervals $\sigma_k = 0.1, 0.2, ..., 0.9$ via gradient descent and use the set of coefficients that $\sigma_k$ is closest to at the current training iteration. Computing the spectral norm and $\sigma_k$ can be done efficiently via the Implicitly Restarted Lanczos Method \cite{Lanczos}.
This can be further sped up by observing that the singular values change slowly over the course of training, allowing us to get away with computing $\sigma_k$ once every 50 iterations.
\cref{ns_table} shows that after 300 epochs $\tilde{W}^\perp$-NS attains comparable accuracies to $\tilde{W}^\perp$, and in some cases even surpasses it while being significantly faster than $W^\perp$. Our $\tilde{W}^\perp$ method and its $\tilde{W}^\perp$-NS variant surpasses the baseline $\tilde{W}\transpose$ in all cases.
For the MLP with a layer width of 512, $W^\perp$-NS has around 10$\times$ smaller memory usage than $W^\perp$.

\begin{table}[t!]
\centering
\caption{Layer 1 variance and bias, and accuracy at 300 epochs for an MLP with layer width of 256 trained on our methods and the baseline $\tilde{W}\transpose$. $\tilde{W}^\perp$, with lower variance and bias than $\tilde{W}\transpose$, reaches the highest accuracy by a wide margin while $\tilde{W}^\text{P}$ (preconditioning), with the lowest bias but highest variance, reaches lower accuracy than $\tilde{W}\transpose$. }
\label{tab:variance_bias_accuracy_256}
\begin{tabularx}{\linewidth}{l C C c C}
\toprule
\textbf{Method} & \textbf{Variance} \eqref{variance} & \textbf{Bias} \eqref{bias_metric} & \textbf{Train Accuracy} & \textbf{Test Accuracy} \\
\midrule
$\tilde{W}^\perp$ ($k=10$)        & $\mathbf{6.33 \times 10^{-6}}$  & $3.63 \times 10^{-5}$  & \textbf{0.433} & \textbf{0.383} \\
$\tilde{W}^\text{P}$    & $1.75 \times 10^{-4}$  & $\mathbf{1.35 \times 10^{-8}}$  & 0.367 & 0.314 \\
$\tilde{W}^\top$           & $7.04 \times 10^{-5}$  & $1.68 \times 10^{-4}$  & 0.375 & 0.316 \\
\bottomrule
\label{preconditioning}
\end{tabularx}
\end{table}

\subsubsection{$\tilde{W}^\text{P}$ Preconditioning Results.}

While the earlier weight perturbation and activation perturbation methods emphasise unbiasedness, the baseline $\tilde{W}\transpose$ and our $\tilde{W}^\perp$ methods emphasise reducing variance but at the cost of incurring bias. 
To gauge the effectiveness of lowering variance versus lowering bias, we compare our $\Tilde{W^\text{P}}$ preconditioning method with $\tilde{W}^\perp$ and the $\tilde{W}\transpose$ baseline in \cref{preconditioning}. 
We see that lower variance correlates with higher accuracy, while the high variance preconditioning method, despite being unbiased, achieves the lowest accuracy. 
This result shows that removing bias at the expense of increasing variance is ineffective, and therefore it is better to have low variance even if it incurs bias. 
Based on the accuracy gains of $\tilde{W}^\perp$ as layer widths increase (\cref{fig:scaling_plot}), we hypothesise that this principle becomes even more relevant for larger networks.

\section{Conclusion}

In this paper, we introduced the $\tilde{W}^\perp$ method to addresses the central problem of high variance in gradient estimation methods.
Our $\tilde{W}^\perp$ method exploits the low-dimensional structure of gradients by orthogonalising the upstream Jacobian matrix, thereby constraining the guessing space and reducing bias and variance. 
Experiments show promising results with $\tilde{W}^\perp$ as network width increases, while the $\tilde{W}^\text{P}$ preconditioning method demonstrates that removing bias at the cost of increased variance is ineffective. 
Furthermore, we greatly speed up $\tilde{W}^\perp$ by using Newton--Schulz iterations for approximate orthogonalisation. 
In future work, these ideas can be expanded to more practical architectures, such as networks with normalisation layers or residual connections, or approaches that better consider the underlying structure of gradients in deep neural networks.

\bibliographystyle{unsrt}
\bibliography{references}

\end{document}